\documentclass[journal]{IEEEtran}
\usepackage{amsmath,amsfonts}
\usepackage[linesnumbered,ruled]{algorithm2e}
\usepackage{amsmath} 
\usepackage{array}
\usepackage[caption=false,font=normalsize,labelfont=sf,textfont=sf]{subfig}
\usepackage{textcomp}
\usepackage{stfloats}
\usepackage{url}
\usepackage{verbatim}
\usepackage{graphicx}
\usepackage{cite}
\usepackage{float}
\usepackage[switch]{lineno} 

\begin{document}
\title{An Evolutionary Network Architecture Search Framework with Adaptive Multimodal Fusion for Hand Gesture Recognition}

\author{
Yizhang Xia,
Shihao Song,
Zhanglu Hou,
Junwen Xu, \\
Juan Zou,
Yuan Liu, Shengxiang Yang,~\textit{Senior Member, IEEE
}
\thanks{Y. Xia, S. Song, Z. Hou, J. Xu,  J. Zou and Y. Liu are with Hunan Engineering Research Center of Intelligent System Optimization and Security, Key Laboratory of Intelligent Computing and Information Processing, Ministry of Education of China, and Key Laboratory of Hunan Province for Internet of Things and Information Security, Xiangtan University, Xiangtan, 411105, Hunan Province, China.
(\emph{Corresponding author: Zhanglu~Hou, e-mail: ahou.amstrong@gmail.com}) }
\thanks{S. Yang  is with the Centre for Computational Intelligence,
School of Computer Science and Informatics, De Montfort University,
Leicester, LE1 9BH, U.K.}

}

\maketitle
\begin{abstract}
Hand gesture recognition (HGR) based on multimodal data has attracted considerable
attention owing to its great potential in applications. Various manually designed multimodal deep networks have performed well in multimodal HGR (MHGR), but most of existing algorithms require a lot of expert experience and time-consuming manual trials. To address these issues, we propose an evolutionary network architecture search framework with the adaptive multimodel fusion (AMF-ENAS). Specifically, we design an encoding space that simultaneously considers fusion positions and ratios of the multimodal data, allowing for the automatic construction of multimodal networks with different architectures through decoding. Additionally, we consider three input streams corresponding to intra-modal surface electromyography (sEMG), intra-modal accelerometer (ACC), and inter-modal sEMG-ACC. To automatically adapt to various datasets, the ENAS framework is designed to automatically search a MHGR network with appropriate fusion positions and ratios. To the best of our knowledge, this is the first time that ENAS has been utilized in MHGR to tackle issues related to the fusion position and ratio of multimodal data. Experimental results demonstrate that AMF-ENAS achieves state-of-the-art performance on the Ninapro DB2, DB3, and DB7 datasets.
\end{abstract}


\section{INTRODUCTION}
Hand gesture recognition (HGR), as a pivotal component of human-computer interaction, has witnessed remarkable advancements in recent years, fueled by the convergence of artificial intelligence and sensing technologies~\cite{1}. Especially, the surface electromyography (sEMG) is a widely-used methodology employing wearable sensors positioned on the skin to capture both motion and physiological signals produced during muscle contractions for HGR~\cite{2, 3}.  sEMG based HGR has attracted considerable attention and it has been widely applied in many fields such as robotics~\cite{4}, virtual reality~\cite{guo2021human} and prosthetic control~\cite{5, 6}. However, its application in real-world scenarios encounters various challenges. These challenges include a reduced signal-to-noise ratio of surface sEMG caused by amputation and the lower discrimination due to the extremely similarity between gestures, ultimately leading to a detrimental decline in performance~\cite{11, 12}.  To address these issues, more and more researchers have focused on exploring the multimodal data composed of sEMG and other sensor data.  One commonly employed multimodal data fusion for HGR involves integrating sEMG and inertial measurement unit (IMU) data. Furthermore, a wealth of research has shown that utilizing deep learning models with multimodal inputs outperform those with single-modal inputs\cite{14, 15, 16}. More specifically, the sEMG, as an intuitive biological signal, can capture subtle nuances in movements, while IMU, comprising accelerometers (ACC), magnetometers (MAG), and gyroscopes (GYR), offers robust data from a physical perspective. The combination of sEMG and IMU data in multimodal approaches has become a popular and effective solution. 

Generally, two critical aspects in HGR are feature extraction and the final recognition accuracy. To acquire sEMG signals, the mainstream approach is to use sparse multi-channel wireless wearable sensors~\cite{7}. The collected data is typically subjected to preprocessing, and during feature extraction, a set of features is employed, incorporating a significant amount of heuristic knowledge~\cite{8}. For example, Lu ~\textit{et al.}~\cite{lu2014hand} achieved successful HGR across various scales by leveraging sEMG and ACC data. The researchers employed a Bayes linear classifier with manually extracted sEMG features to discern small-scale gestures, where they utilized a dynamic wrap algorithm incorporating fused manual features of both sEMG and ACC for the classification of large-scale gestures. However, the classification of gesture scales in their work is a labor-intensive process, and the multimodal features were directly input into the classifier after splicing. Besides, an increasing body of research is adopting an end-to-end multimodal fusion approach using deep-learning technology~\cite{yang2021dynamic, wei2019surface}. These approach significantly facilitate the development of HGR systems based on sEMG and inertial IMU data. However, many of these methods employ single-scale convolution operations for extracting deep features, and this often leads to an incomplete representation of the multiscale properties inherent in multimodal data and fails to capture the hierarchical relationships within deep features~\cite{11}.
In addition to the challenges mentioned above in multimodel HGR (MHGR), the majority of current methods still require a substantial amount of expert knowledge and time for manually designing a suitable network architecture and feature extraction.
To  efficiently improve the accuracy of HGR, this paper proposes an evolutionary network architecture search with the adaptive multimodal fusion (AMF-ENAS) to achieve the automatic construction of multimadal networks. Meanwhile, a multimodal fusion strategy that simultaneously employs adaptive adjustments to the fusion position and ratio to enhance the performance of a trained multimodal network in the face of varying characteristics of different data. The main contributions of AMF-ENAS are summarized as follows:
\begin{itemize}
    \item \textbf{Evolutionary multimodal  network architecture search framework:} We have proposed a block-based evolutionary multimodal  network architecture search framework. This framework achieves better performance than manually designed conventional multimodal networks in MHGR. We conducted comprehensive evaluations of our proposed method on multiple MHGR databases to verify the performance of our evolutionary multimodal  network architecture search framework. Experimental results show that our proposed framework achieves higher accuracy of HGR on multimodal data streams than manually designed multimodal deep networks.
    \item \textbf{A novel multimodal fusion strategy:} We propose a novel multimodal fusion strategy that considers both the positioning of different data fusion nodes in the model and the fusion ratios between data from different branches. We fully consider the features of data in both shallow and deep layers of the model, as well as the importance of different data streams.
    \item \textbf{An encoding strategy for adapting multimodal data:} To adapt to multimodal data, we reshape the encoding space, delineating it into three functional components: fusion points, fusion ratios, and block selection for control. Specifically, each layer of the network architecture consists of customizable blocks, with an additional specific searchable space incorporated on top of the fixed structure at each layer. Based on the encoding strategy, we can decode various encodings into distinct deep neural network architectures. Through the iterative application of genetic algorithms, this method can quickly adapt to different multimodal databases to search suitable multimodal deep networks.
\end{itemize}

\section{BACKGROUND AND RELATED WORKS}
The current mainstream approach in MHGR related to sEMG signals uses deep learning methods.Deep learning algorithms are usually crafted with multi-layer data representation architectures. These frameworks initially extract low-level attributes from the data at the early layers, extracting progressively high-level features as they advance toward the final layer. The strategic design of these deep networks revolves around maximizing feature extraction and utilizing these identifications to execute tasks. Nevertheless, reliance on manually curated network architectures built on seasoned expertise may only sometimes yield optimal solutions for the HGR task. Recently, with the ascent of Evolutionary Neural Architecture Search (ENAS) within the realm of automated deep neural networks, ENAS has been regarded as a promising approach to counter the challenges posed by manual network design. Subsequent sections will offer detailed insights into hand gesture recognition and ENAS.

\subsection{Hand Gesture Recognition}
Emerging as a prominent technology within human-computer interaction, the HGR has been extensively applied and developed across many fields. HGR can take various forms, contingent on the utilized sensing technology, encompassing data gloves, visual sensing, and a range of wearable devices built upon sEMG and ultrasonic signals. Our article primarily concentrates on the technology for gesture recognition built on sEMG signals. The prevailing methods for constructing classifiers include traditional machine-learning techniques and sophisticated deep-learning methods~\cite{2_0_zheng2022surface}. The traditional machine learning approaches consist of linear discriminant analysis (LDA)~\cite{2_1_fougner2011resolving}, support vector machine (SVM)~\cite{2_2_chen2017exploring}, neural network models~\cite{2_3_englehart1999classification}, decision trees~\cite{2_4_wu2016wearable}, hidden Markov models \cite{2_5_zhang2011framework}, K-nearest Neighbors~\cite{2_6_kim2008bi}, Gaussian mixture models~\cite{2_7_huang2005gaussian}, kernel regularized least squares methods~\cite{2_8_gijsberts2014movement}, Naive Bayes classifiers~\cite{2_9_amma2015advancing}, and locally weighted projection regression~\cite{2_10_castellini2009surface}. LDA is most typically employed as a general classifier for gesture recognition. Despite the achievements of traditional machine learning algorithms and their relative simplicity in understanding and implementation, they often need to improve compared to deep learning methods as the data escalates in complexity and volume.

Deep learning technology has demonstrated outstanding performance across numerous domains in recent years. Concurrently, several studies have successfully applied deep learning models such as convolutional neural networks (CNN) and recurrent neural networks (RNN) to sEMG, thereby achieving commendable results in gesture recognition \cite{2_11_li2021gesture}. A user-adaptive CNN model, conceived by Park and Lee ~\cite{2_12_park2016movement}, is acknowledged as the inaugural deep learning-based framework applied to sEMG signals for data classification in the Ninapro database. Simultaneously, a prevailing belief among certain studies is that surface electromyography fundamentally manifests as time-series signals, thereby making it considerably more apt for recurrent neural networks (RNN) that process time series or other sequence information~\cite{2_13_hu2018novel}. A model combining a long short-term memory  network and a multi-layer perceptron for feature learning and sEMG signal classification was proposed \cite{2_14_he2018surface}, which enhanced motion classification accuracy by constructing a feature space that accommodates both the dynamic and static information of sEMG signals. Considering the temporal correlation, Temporal Convolutional Networks (TCN) have also found application within the research work focused on gesture recognition based on surface electromyography. In their study, Hauser ~\textit{et al.}~\cite{2_15_betthauser2019stable} utilized the historical data of sEMG signals to discern temporal features. Recent research has inclined towards the development of more fortified fusion networks by integrating additional input streams. A method known as multi-stream CNN, implemented by Wei ~\textit{et al.}~\cite{15}for gesture recognition, was confirmed to outshine both random forest and elementary CNN. Duan~\textit{et al.}~\cite{16} proposed a novel and practical hybrid fusion model with multi-scale attention and metric learning, achieving the best performance to date in natural gesture classification using sEMG and ACC data from three sub-databases of NinaPro~\cite{atzori2012building}. Despite the promising performance of manual networks in hand gesture recognition, they necessitate significant expert experience and many human trials for network architecture modification throughout the construction process. To address this issue, our approach involves implementing an evolutionary neural network architecture search method to autonomously construct multimodal hand recognition networks tailored to distinct datasets.
\begin{figure*}
    \centering
    \includegraphics[scale = 0.55]{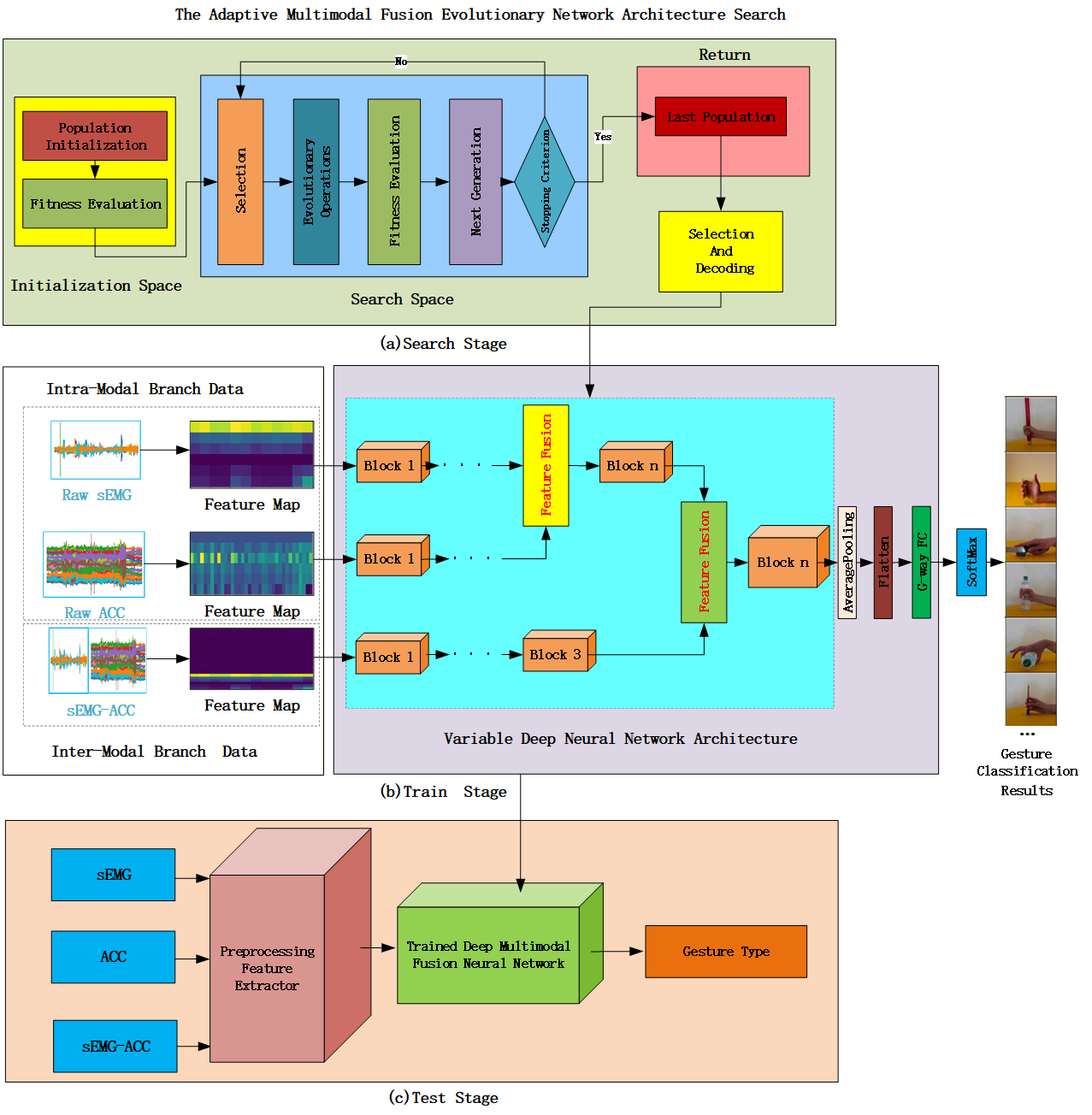}
    \caption{Illustration of the proposed framework for an adaptive multimodal evolutionary network architecture search. (a) Search for suitable multi-modal deep network architecture. (b) Train the searched network. (c) Test the final performance after fine-tuning on the trained network.  “Block1” and “Block3” represent fixed block types, while “Blockn” represents an undecided block type.}
    \label{fig1}
\end{figure*}

\subsection{Evolutionary Network Architecture Search}
The crux of ENAS revolves around the utilization of Evolutionary Computation (EC) technology to tackle Neural Architecture Search (NAS) issues. EC adopts the principles of natural evolution or collective behavior and has found wide application in confronting optimization problems~\cite{hou2020reformulating, xu2024cluster}. Well-known methods like Particle Swarm Optimization~\cite{30}, Genetic Algorithm (GA) ~\cite{29}, and Ant Colony Optimization (ACO) ~\cite{31} are in ample practice today. The inception of evolutionary computational techniques for automatically searching neural network architectures and weights is termed NeuroEvolution ~\cite{32}. The distinguishing attribute between ENAS and NeuroEvolution is ENAS's focus on architecture search, deemed suitable for deep neural networks in most instances. The influx of ENAS workloads commonly gets ascribed to Google's proposed LargeEvo algorithm ~\cite{33}, which harnesses GA to seek the optimal structure of the CNN. The effectiveness of this algorithm is veritable via an encompassing series of experimental outcomes. Novel techniques, such as search strategies based on reinforcement learning, have been introduced within ENAS to heighten the efficiency and effectiveness of the search. Chen~\textit{et al.}~\cite{34} propagated a reinforcement learning-guided evolutionary mutation technique for NAS, executing the search for resilient network architectures on the CIFAR-10 dataset ~\cite{thakkar2018batch}. The proposed architecture showcases competitive results on CIFAR-10 and elevates to unprecedented accuracy levels when applied to the ImageNet dataset~\cite{kornblith2019better}. As the influence of skip connections takes a more tangible turn, a separate research~\cite{35}  proposed a genetic algorithm that searches the space of all networks between a standard feed-forward network and the corresponding DenseNet.In the process of evolutionary neural network architecture search, the search space outlines the set of all possible neural network architectures that the search could explore, while the encoding space details how to present these architectures in a manner understandable to computers. Thus, a proficiently constructed encoding space is critical for the ENAS algorithm, as it includes the search space, influences the potential uncovering of network architecture, and ultimately boosts the recognition of high-level networks. 
\begin{figure*}
    \centering
    \includegraphics[scale = 0.6]{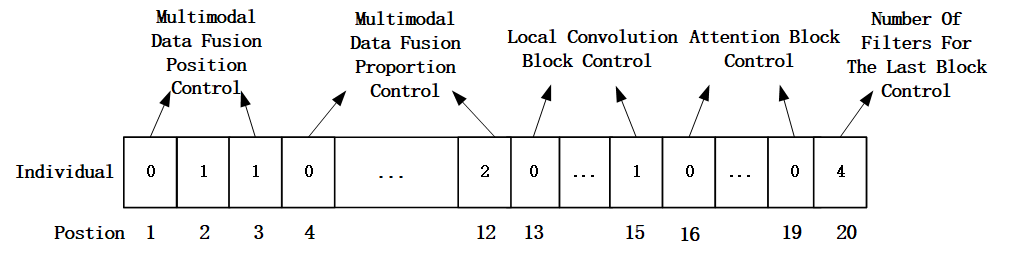}
    \caption{Illustration of the proposed encoding scheme with fixed-length encoding, which each segment of the encoding corresponds to a specific functionality.}
    \label{fig2}
\end{figure*}

Generally, depending on the distinct units implemented, the encoding space can be classified into three types: layer-based encoding space ~\cite{37, 38}, block-based encoding space ~\cite{39}, and cell-based encoding space ~\cite{40}. In the layer-based encoding space, basic layer structures of neural networks, such as convolutional and pooling layers, are typically the foundational units. However, given deep neural networks (DNN) typically comprise many layers, the layer-based search space can be expansive. This necessitates encoding a large volume of information, making the search for the ideal DNN time-intensive. In addition, creating high-functioning DNNs using a limited number of basic layer structures can be challenging, which makes block-based coding spaces a preferable alternative. For instance, a residual block is a block with a specific topology. Searching for efficient architectures in the block-based coding space is more straightforward than in the layer-based coding space, as these blocks have been previously verified to perform well~\cite{38}.Moreover, constructing neural networks with block structures requires fewer coding parameters, reducing coding complexity and search time. Although cell-based and block-based coding spaces are connected, the former can be considered a specific case of the latter. Hence, upon thorough consideration, we opted for a block-based coding space. During the coding design process, we considered the position of data fusion and the impact of the data fusion ratio. Based on these considerations, we propose a block-based evolutionary neural network architecture search framework suitable for multimodal gesture recognition.


\section{METHODOLOGY}

\subsection{An Evolutionary Network Architecture Search Framework with Adaptive Multimodal Fusion}

Considering varying importance among different data modalities in HGR, we propose AMF-ENAS to simultaneously explore the fusion positions and ratios of different data in the network, aiming to address the limitations of manually designed fusion schemes.The AMF-ENAS framework, as illustrated in Fig.~\ref{fig1}, consists of three stages: search, train, and test.
During the search stage, we execute both rough and transfer searches. A rough search is employed primarily to obtain an initial population of high-quality individuals. This involves extracting a portion of the data from all sub-datasets and conducting the search on the combined dataset. Thereafter, this population serves as the starting point for the transfer search, as delineated in algorithm~\ref{algo1}. Ultimately, the aim is to obtain the ideal fusion network for each sub-dataset through the transfer search. Built on the population gleaned from the rough search, we perform individual searches on each sub-dataset to gather deep neural network architectures specific to the characteristics of each sub-dataset.During the subsequent train stage, we further refine and adapt the best individual, referring to the optimal deep neural network acquired during the search stage.  
Lastly, in the testing phase, we execute adaptive fine-tuning to customise the network more precisely towards the specific requirements of the target task, which further augments its performance. Following this, we assess the network's performance. Detailed descriptions of the primary components of the AMF-ENAS framework are provided below.

\begin{algorithm}
       \label{algo1}
        \SetAlgoLined 
        \renewcommand{\thealgocf}{1} 
        \caption{Framework of AMF-ENAS} 
        \KwIn{population size S. } 
        \KwOut  {Best individual.}
         $P_0 \gets $ Initialize the population in the search space with multimodal fusion encoding strategy\; 
         
        $t \gets 0$\; 
       \While{termination criterion is not satisfied}{
        Evaluate the fitness of individuals in Pt\;
        $Q_t \gets $ Generate offspring from P using genetic operators\;
        $P_{t+1} \gets $ Environmental selection from $Q_t$ by using the proposed strategy\;
        $t \gets t +1 $
	        }
         
         $P_r \gets $ Select the best S/2 individuals from $p_t$\;
         $P_n \gets $ Initialize S/2 individuals with multimodal fusion encoding strategy\;
         $P_t \gets $ $P_r \cup P_n$ \;
         
        \While{termination criterion is not satisfied}{
        $Q_t \gets $ Generate offspring with the designed genetic operators from $P_t$\;
        Evaluate the fitness level of newly generated individuals post-transfer\;
        $P_{t+1} \gets $ Select the best S individuals from $Q_t$\;
	        }
    $C \gets $ Select the best individual from $P_t$ and decode it to generate a corresponding multimodal deep neural network\;
	return C.
\end{algorithm}
\subsubsection{Search stage}
The search process is composed of two primary components: the rough search and the transfer search. The ultimate objective is to secure high-quality individuals, which in this context are multimodal DNNs fitting to the sub-datasets. The rough search stage principally targets acquiring a superior population on a combined external dataset encompassing data from all datasets. This population is then utilised as the onset population for the transfer stage on the sub-datasets. We have contrived an encoding methodology for the multimodal networks, wherein each network equates to an individual in the evolutionary algorithm. Initially, a population consisting of a certain quantity of individuals is established. Every individual is then decoded to generate a multimodal neural network. The performance of each network on the mixed dataset is assessed, and the validation loss is adopted as the fitness value for each individual. The roulette wheel selection method~\cite{yu2016improved} picks superior individuals based on their relevant fitness values. Crossover and mutation operations are carried out to produce the subsequent generation. After numerous iterations, the individual with the top fitness value is selected as the ultimate network architecture. The superior individuals obtained from this process serve as the inception population for the subsequent stage. The second component, the transfer search, primarily varies from the initial part in terms of the dataset utilised. The initial population, derived from a rough search on the mixed dataset, is employed as the starting population for the transfer search. The search continues on the sub-datasets to find individuals adapted to each sub-dataset.
\begin{figure}
    \centering
    \includegraphics[scale = 0.6]{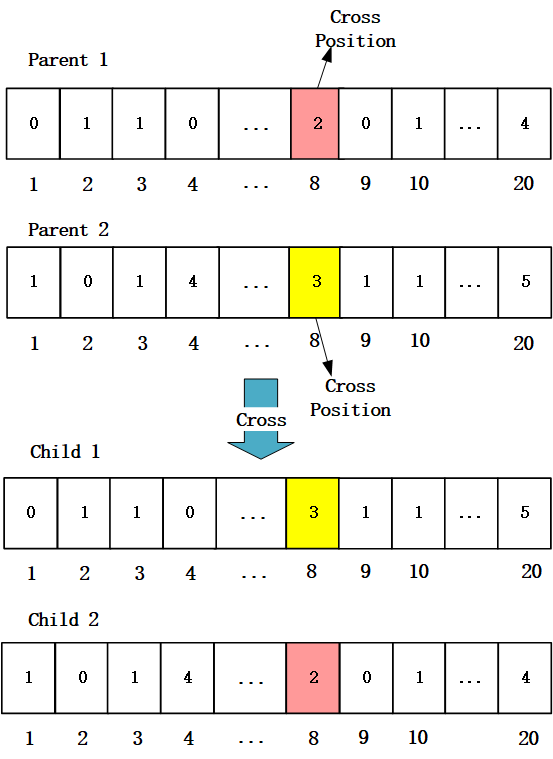}
    \caption{The process of producing offspring through crossover in the evolutionary process.}
    \label{fig3}
\end{figure}

\textbf{Population initialization:} Before initializing the population, it is essential to restrict the size of the networks to prevent the search from yielding excessively large or small networks. Drawing upon past experiences with manually designed networks, we limit the depth of the network by restricting the number of blocks to 4-6. Each individual in the population has a length of 20, and the encoding scheme is depicted in Fig.~\ref{fig2}. According to the encoding scheme, positions [1-3] correspond to the fusion point locations. The fusion of multimodal data is broken down into pairwise fusion. In the case of three data streams, there are two fusion points. The first two encodings determine the locations of these two fusion points, after which blocks are added. The third encoding specifies which two data streams are fused initially. Positions [4-12] control the proportion of filters in each block. Positions [13-15] decide whether to add a local convolutional block after the fusion point, the position following which the block is added, and the number of filters in that block. Positions [16-19] govern the attention block and insertion point. The final position controls the number of filters in the last block. Each encoding has a candidate range. The initial population is randomly generated based on the encoding above scheme, with a fixed number of individuals.

\begin{figure}
    \centering
    \includegraphics[scale = 0.6]{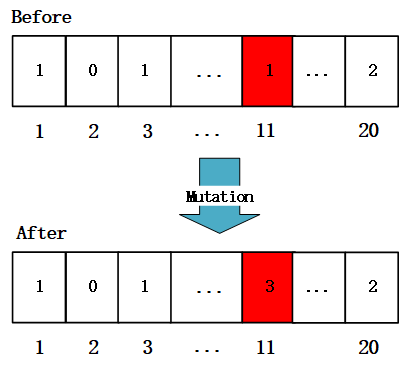}
    \caption{Illustration of the process of mutation during the evolutionary process.}
    \label{fig4}
\end{figure}
\begin{figure*}
    \centering
    \includegraphics[scale = 0.5]{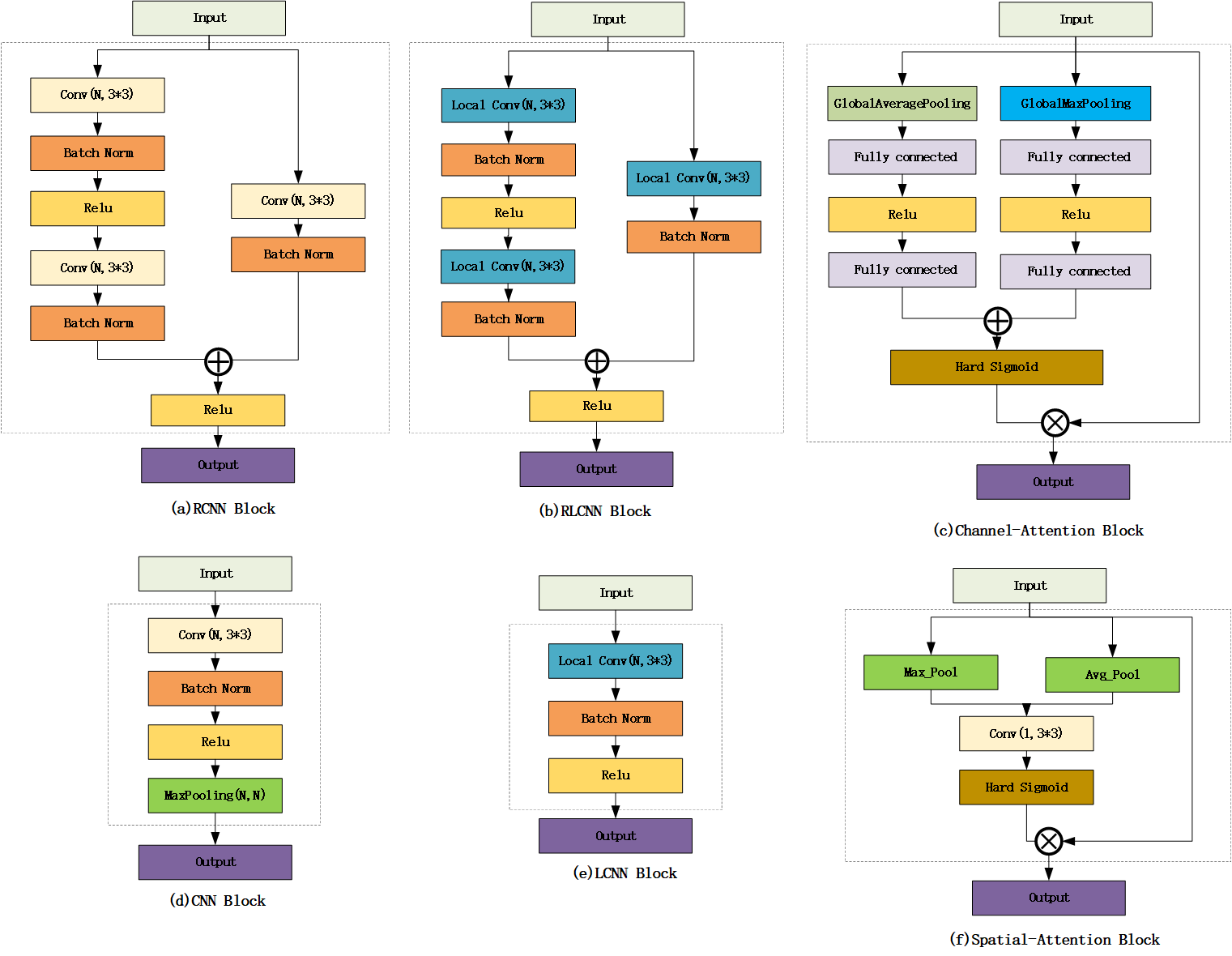}
    \caption{There are six different block structures, with an illustration of the internal structure of each block. (a) Residual block based on ordinary convolution. (b) Residual block based on local convolution. (c) Channel-Attention block (d) Ordinary convolution block. (e) Local convolution block. (f) Spatial-Attention block.}
    \label{fig5}
\end{figure*}

\textbf{Environment selection:} An environmental selection process is implemented to choose individuals for the forthcoming generation. However, it is essential to note that not all offspring generated will be incorporated into the next generation. Instead, offspring are organized based on their fitness evaluations, and only the top-performing individuals are chosen to move forward. Each individual’s decoded network is assessed by training it on the corresponding dataset during the fitness evaluation phase. Considering the time-consuming nature of training a neural network, the networks are only trained for a few epochs during this phase. The final loss on the validation set is utilized as the fitness score for each individual.

\textbf{Generation of offspring:} The evolutionary algorithm produces a novel offspring population from a pre-existing population via crossover and mutation operations. The crossover operation entails a local search, whereas the mutation operation involves a global search. Both of them collectively strive to yield superior-performing offspring. Based on the fitness of the preceding generation, two individuals are selected as parents for the subsequent generation using the roulette wheel selection method. Our approach involves a single-point crossover operation, as depicted in Fig.~\ref{fig3}. This operator facilitates crossover between two individuals, and a crossover point is elected for each individual to exchange their encoded information based on the crossover rate. The single-point mutation operator utilized in this study is portrayed in Fig.~\ref{fig4}. Similarly, if a mutation occurs based on the mutation rate, a mutation point is selected, and the encoding at that point is randomly generated within the specified range. Both crossover and mutation are executed to augment diversity and breed the best-fitting individuals.
\subsubsection{Train stage}
Following the initial stage of rough and transfer search, the optimal individual, which constitutes the multimodal DNN architecture adaptable to each sub-dataset, is acquired. This optimal network is individually trained for each sub-dataset during the training stage.
\subsubsection{Test stage}
Nevertheless, differences in distribution between the training and testing data may influence gesture recognition performance amongst different subjects. To mitigate this issue, a certain degree of fine-tuning \cite{41} is enacted during the testing stage that suits each sub-dataset. This process of fine-tuning assists in closing the gap between the training and testing distributions, thereby improving the network's performance on each sub-dataset.

\subsection{A Novel Multimodal Fusion Strategy}
Not only inspired by traditional fusion methods like early and late fusion, we also pinpoint research gaps in multimodal gesture recognition. No existing research considers the fusion ratio of multimodal data, which is vital at the fusion junctures, given its importance to various data. To tackle this issue, we propose an innovative multimodal fusion strategy in this project. Our solution contemplates the positions of various data fusion nodes within the model and the fusion ratios between differing branch data. The quantity of filters in a convolutional block dictates the volume and ratio of data fused when uniting branches. In this experiment, we considered a total of 10 distinct candidate filters.

\subsection{An Encoding Strategy for Adapting Multimodal Data}
To accommodate multimodal data, we have restructured the coding space of AM-ENAS. Now, each layer within the neural network comprises customizable blocks, and we've included a specific searchable space based on the network's fixed layer structure. The block selection relies on successful structures from previous studies that utilized surface electromyography signals for gesture recognition. In this experiment, we have adopted the local convolutional structure from the MV-CNN network, which showcased exceptional performance in the experiment led by Wei et al. \cite{15}. Additionally, we have incorporated the attention mechanism and merged it with the confirmed efficacy of residual structures observed in diverse studies. Our methodology includes six different types of blocks: regular convolutional block, residual convolutional block, local convolutional block, local residual convolutional block, channel attention block, and spatial attention block \cite{woo2018cbam}. The architectures of these blocks are illustrated in Fig.~\ref{fig5}.

\section{EXPERIMENTAL RESULTS}
\subsection{Datasets}
Our evaluation is conducted on three sub-datasets (DB2, DB3, DB7) from the Ninapro database containing synchronized SEMG and ACC signals. The Ninapro database is a publicly available multimodal database for researching and validating machine learning-based human, robotic, and prosthetic control systems. It is a valuable resource in electromyographic gesture recognition, providing abundant SEMG data for studying and developing myoelectric hand prostheses and other muscle signal applications. The specifications of the three selected sub-datasets are presented in Table~\ref{table1}.

\begin{table}
\captionsetup{font={scriptsize}}
\captionsetup{width=1.0\linewidth}
\caption{The specifications of the datasets evaluated in this research}
\label{table1}
\resizebox{1.0\columnwidth}{4.5cm}{%
\fontfamily{ptm}\selectfont 
\fontsize{10}{14}\selectfont 
\begin{tabular}{cccc}
\hline
Dataset Name                                                                  & Ninapro DB2\cite{42} & Ninapro DB3\cite{42} & Nianapro DB7\cite{LDA} \\ \hline
\begin{tabular}[c]{@{}c@{}}Total number \\ of gestures\end{tabular}           & 50          & 50          & 41           \\\hline
\begin{tabular}[c]{@{}c@{}}Number of gestures\\ to be classified\end{tabular} & 50          & 50          & 41           \\\hline
Intact subjects                                                               & 40          & 0           & 20           \\\hline
Amputed subjects                                                              & 0           & 11          & 2            \\\hline
\begin{tabular}[c]{@{}c@{}}Number of\\ sEMG channels\end{tabular}             & 12          & 12          & 12           \\\hline
\begin{tabular}[c]{@{}c@{}}Number of\\ Accelerometer channels\end{tabular}    & 36          & 36          & 36           \\\hline
\begin{tabular}[c]{@{}c@{}}Number of \\ Gyroscope channels\end{tabular}       & 0           & 0           & 36           \\\hline
\begin{tabular}[c]{@{}c@{}}Number of \\ Magnetometer channels\end{tabular}    & 0           & 0           & 36           \\\hline
Number of trials                                                              & 6           & 6           & 6            \\\hline
Trials for training                                                           & 1,3,4,6     & 1,3,4,6     & 1,3,4,6      \\\hline
Trials for testing                                                            & 2,5         & 2,5         & 2,5          \\\hline
Sampling rate                                                                 & 2000Hz      & 2000Hz      & 2000Hz       \\ \hline
\end{tabular}%
}
\end{table}
Ninapro DB2 \cite{42} provides sEMG and ACC signals from 40 subjects. Each subject was asked to perform three sets of actions, which consisted of 17 bare finger and wrist movements (Exercise B), 23 grasping and functional movements (Exercise C), and 9 power grasp movements (Exercise D), resulting in a total of 50 gestures. Each gesture was attempted six times, each action lasting 5 seconds and alternating with a rest posture lasting 3 seconds. For signal acquisition, 12 Trigno wireless electrodes from Delsys, Inc. (www.delsys.com) were utilized. The EMG signals were sampled at a frequency of 2 kHz. These electrodes also integrated a 3-axis accelerometer, which was sampled at a rate of 148 Hz. All data streams were super-sampled to the highest sampling frequency of 2 kHz to ensure consistency.
\begin{table}
\captionsetup{font={scriptsize}}
\captionsetup{width=1.0\linewidth}
\caption{The specifications of handcrafted features for sEMG and ACC signals}
\label{table2}
\resizebox{1.0\columnwidth}{2cm}{%
\fontfamily{ptm}\selectfont 
\fontsize{10}{14}\selectfont 
\begin{tabular}{cccc}
\hline
\multicolumn{1}{r}{sEMG Features} & \multicolumn{1}{l}{Feature Map Size} & \multicolumn{1}{l}{ACC Features} & \multicolumn{1}{l}{Feature Map Size} \\ \hline
IEMG                              & 1×12                                 & MEAN                             & 1×36                                 \\
WL                                & 1×12                                 & VAR                              & 1×36                                 \\
VAR                               & 1×12                                 & RMS                              & 1×36                                 \\
ZC                                & 1×12                                 & WL                               & 1×36                                 \\
SSC                               & 1×12                                 & MAV                              & 1×36                                 \\
WAMP                              & 1×12                                 & MAVS                             & 1×36                                 \\ \hline
Total                             & 6×12                                 & Total                            & 6×36                                 \\ \hline
\end{tabular}%
}
\end{table}

\begin{table*}
\captionsetup{font={small}}
\captionsetup{width=1.0\linewidth}
\caption{The experimental results of each method in the original paper on Ninapro datasets. The results demonstrate that the AMF-ENAS achieves the highest performance.}
\label{table4}
\resizebox{2.0\columnwidth}{0.5\columnwidth}{%
\fontfamily{ptm}\selectfont 
\fontsize{9}{14}\selectfont 
\begin{tabular}{cccccc}
\hline
\multicolumn{1}{l}{Method} & \multicolumn{1}{l}{Dataset} & \multicolumn{1}{l}{Number of gesture to be classified} & Modalities & Window length & Accurancy        \\ \hline
KRLS \cite{gijsberts2014movement}                      & Ninapro DB2                  & 40                                                     & sEMG-ACC   & 400ms         & 82.49\%          \\
Improved KRLS \cite{atzori2014classification}             & Ninapro DB2                  & 40                                                     & sEMG-ACC   & 400ms         & 92.10\%          \\
MV-CNN     \cite{15}                 & Ninapro DB2                  & 50                                                     & sEMG-IMU   & 200ms         & 94.40\%          \\
HyFusion    \cite{16}                 & Ninapro DB2                  & 50                                                     & sEMG-ACC   & 200ms         & 94.73\%          \\
\textbf{AMF-ENAS}          & Ninapro DB2                  & 50                                                     & sEMG-ACC   & 200ms         & \textbf{95.15\%} \\ \hline
Improved KRLS   \cite{atzori2014classification}            & Ninapro DB3                  & 40                                                     & sEMG-ACC   & 400ms         & 88.90\%          \\ 
SVM  \cite{liu2016continuous}                      & Ninapro DB3                  & 40                                                     & sEMG-ACC   & 250ms         & 88.72\%          \\
MV-CNN       \cite{15}                & Ninapro DB3                  & 50                                                     & sEMG-IMU   & 200ms         & 87.06\%          \\
HyFusion      \cite{16}               & Ninapro DB3                  & 50                                                     & sEMG-ACC   & 200ms         & 89.60\%          \\
\textbf{AMF-ENAS}          & Ninapro DB3                  & 50                                                     & sEMG-ACC   & 200ms         & \textbf{92.50\%} \\ \hline
LDA    \cite{LDA}          & Ninapro DB7                  & 40                                                     & sEMG-ACC   & 256ms         & 82.70\%          \\
MV-CNN     \cite{15}           & Ninapro DB7                  & 41                                                     & sEMG-IMU   & 200ms         & 94.54\%          \\
HyFusion   \cite{16}                  & Ninapro DB7                  & 41                                                     & sEMG-ACC   & 200ms         & 96.44\%          \\ 
\textbf{AMF-ENAS}          & Ninapro DB7                  & 41                                                     & sEMG-ACC   & 200ms         & \textbf{97.19\%} \\ \hline
\end{tabular}%
}
\end{table*}
Ninapro DB3 \cite{42} provides sEMG and ACC signals from 11 subjects who have undergone radial artery amputation. These subjects were instructed to perform the same actions as in DB2, which includes the 50 gestures and the rest posture. The data collection conditions, including the sampling rate, were consistent with DB2. As mentioned by the database authors \cite{42}, it is worth noting that two amputated subjects (subject 7 and 8) had fewer electrodes than the other 10 subjects due to space limitations. Additionally, three amputated subjects (subject 1, 3, and 10) had to interrupt the experiment before completion due to fatigue or pain. Therefore, similar to the approach in \cite{16}, only subjects who used the complete number of electrodes and completed all actions were considered for training and testing in this study.

Ninapro DB7 \cite{LDA} provides sEMG and IMU signals (including ACC, MAG, and GYR) from 20 intact subjects and 2 subjects with radial artery amputation. The signal acquisition protocol is identical to that in Ninapro DB2 and DB3. Each subject performed two sets of actions, consisting of 17 bare finger and wrist movements (Exercise B), 23 grasping and functional movements (Exercise C), and a rest posture, resulting in 41 gestures. Similarly, this study did not consider data from amputated subjects with insufficient channel numbers following the approach in \cite{16}.

It should be noted that only sEMG signal data and ACC signal data are used in any of the datasets mentioned in this experiment.
\subsection{Experimental Setup}
All experiments in this study were conducted on a workstation equipped with an NVIDIA GeForce GTX 3090 GPU. The proposed model was implemented using TensorFlow and Keras. We followed the classical electromyographic (EMG) signal control process \cite{hudgins1993new}.During the rough search stage, we performed 20 generations of searches. Due to the more significant amount of data in the transfer search stage, we conducted 5 generations of search on each sub-dataset. In the final training stage, each network underwent 25 training epochs.The Adam optimizer with adaptive moment estimation was utilized for the training process. The learning rate was initialized as 0.001 and dynamically decreased by dividing it by 10 after the 12th and 20th epochs.The overall experiment involved signal preprocessing, feature extraction, and classification. For the classification stage, we trained the network obtained through ENAS search. To address the potential impact of distribution differences between training and testing data on gesture recognition performance, we followed the approach of Wei et al. \cite{15} by adapting with a small amount of testing data. This adaptation helped mitigate the adverse effects caused by distribution differences between the training and testing data.
\subsubsection{Data preprocessing}In the data preprocessing stage, we extract the sEMG and ACC data streams from each subject and divide them into training and testing sets based on gesture labels and repetition experiment labels. Additionally, we normalize the data to ensure consistent scaling. Next, in the training set, we process the boundaries of each repeated action and rest posture by removing 10\%  of the blurry data at the boundaries, and the test set will not be processed. To segment the processed data, we utilize a sliding window approach with a window length of 200ms and a step size of 10ms. Previous studies have shown that the window length for electromyographic (EMG) control applications should generally be within 300ms \cite{krasoulis2019multi}. Additionally, previous research has demonstrated that a 200-ms window is sufficient to meet the control requirements of EMG signals.
For the segmented data, we perform feature extraction based on the specific requirements of the classification algorithm or model used in the study.

\subsubsection{Feature extraction} This article utilizes classic manual sEMG features, including time domain and frequency domain features. These features consist of electromyography integral (IEMG), variance (VAR), Verizon amplitude (WAMP), waveform length (WL), slope sign change (SSC), zero crossing (ZC), root mean square (RMS), mean absolute value (MAV), and MAV slope (MAVS).We extracted six features for sEMG and ACC based on previous research. The specifications of the manual features used in this study are shown in Table~\ref{table2}.Similar to the approach of Duan et al. \cite{16}, we reshaped the ACC hand-made feature map from a size of 6×36 to 18×12. We then concatenated it with the sEMG hand-made feature map of size 6×12. As a result, we obtained a hand-crafted feature map of size 24×12, which will be used as the input data for the third stream.

\subsubsection{Evaluation metric}The evaluation method used in this article is consistent with that of previous studies \cite{15,16,42}. From each subject’s six repeated actions, the trial data with identification numbers 1, 3, 4, and 6 are used for training, while the trial data with identification numbers 2 and 5 are used for testing.The experiments in the article were conducted using the subjects' internal data and externally extracted datasets. The final gesture recognition accuracy was calculated using the evaluation index shown in Equation~\ref{e1} by averaging the accuracy of all subjects.
\begin{equation}
\label{e1}
\scriptsize{\mathrm{\normalsize{\mathbf{Accuracy= \frac{Number\, of \, correctly \, classified \, samples}{Total\, number\, of \, samples} * 100\%  }}
}}
\end{equation}
\subsection{Comparison with NAMF-ENAS and Manually Designed Networks}
\begin{figure*}
    \centering
    \includegraphics[scale = 0.6]{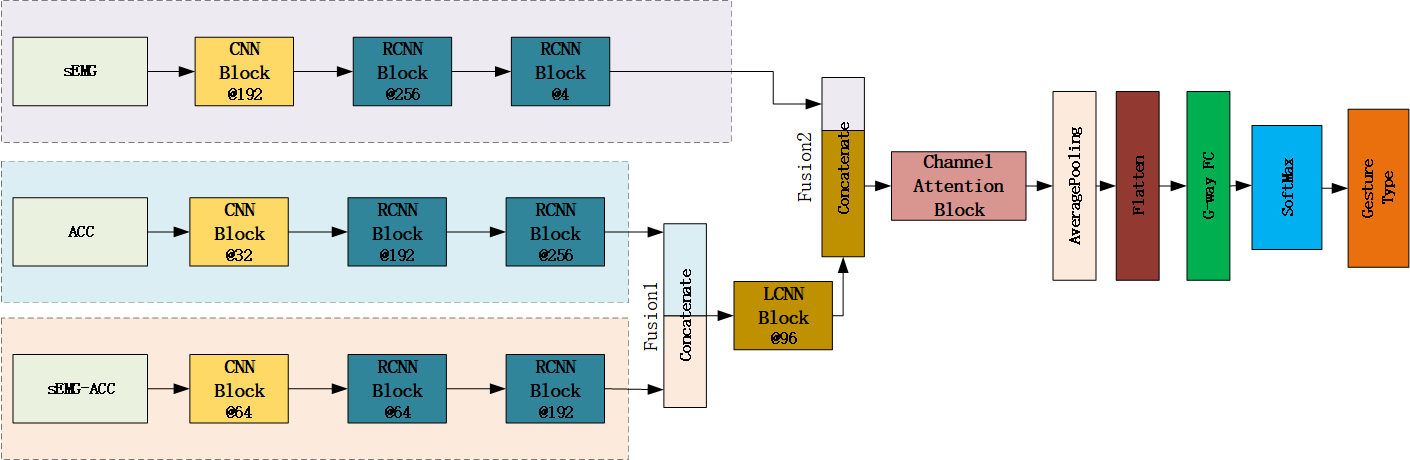}
    \caption{The network architecture searched on a sub-dataset of DB2 is shown in the figure. The internal structure of the Block used in the network is shown in Figure 5. The number after the symbol (@) denotes the number of convolution filters in the Block. Fusion1 and Fusion2 represent two fusion point positions, and the convolution filter of the previous layer controls their fusion ratio.}
    \label{fig6}
\end{figure*}

In this section, we compare the effectiveness of our proposed AMF-ENAS approach with two other experiments: 1) NAMF-ENAS, which does not consider the fusion ratio of multimodal data, and 2) an experiment using an artificial neural network constructed with the same basic building blocks and input data as our approach. The experiments are conducted on the Ninapro DB2, DB3, and DB7 databases using the same hyperparameter settings. The network architecture that we used AMF-ENAS to search on one of the sub-datasets is illustrated in Figure~\ref{fig6}. The results of the comparative experiments are presented in Table~\ref{table3}.
The experimental results demonstrate the superiority of the AMF-ENAS approach, which considers both fusion points and ratios, over the approach that only considers fusion points and the artificial neural network approach. These results highlight the effectiveness of our multimodal AMF-ENAS method for gesture recognition using multimodal SEMG and ACC data streams.
\subsection{Comparison with The State-of-the-art Gesture Recognition Methods}
In this section, we compare the performance of our proposed AMF-ENAS model with other representative artificial neural networks that utilize both SEMG and ACC data. We evaluate the performance of three databases: Ninapro DB2, DB3, and DB7. The comparison includes traditional methods and deep learning algorithms, such as Kernel Regularized Least Squares (KRLS) \cite{gijsberts2014movement}, Improved KRLS \cite{atzori2014classification}, Support Vector Machine (SVM) \cite{liu2016continuous}, Linear Discriminant Analysis (LDA) \cite{LDA}, Multi-View CNN (MV-CNN) \cite{15}, and the state-of-the-art work HyFusion \cite{16}.

In our comparison, we primarily focus on multimodal studies and compare the experimental results with other studies conducted on Ninapro DB2, DB3, and DB7. It is worth noting that some studies only used a subset of gestures to validate their methods, while we utilized all gestures in the dataset, following the approach of \cite{15,16}.As shown in Table~\ref{table4}, our proposed AMF-ENAS achieves average recognition accuracies of 95.15\%, 92.50\%, and 97.19\% for multimodal gesture recognition on Ninapro DB2, DB3, and DB7, respectively. These results demonstrate that our approach outperforms existing artificial multimodal deep neural networks regarding recognition accuracy.

\begin{table}
\captionsetup{font={footnotesize}}
\captionsetup{width=1.0\linewidth}
\caption{The ablation experiments are compared with the average accuracy of the integration ratio and the average accuracy of the pure artificial network. The results demonstrate that the AMF-ENAS achieves the highest performance.}
\label{table3}
\resizebox{1.0\columnwidth}{2cm}{%
\fontfamily{ptm}\selectfont 
\fontsize{11}{14}\selectfont 
\begin{tabular}{lllc}
\hline
Dataset             & \multicolumn{1}{c}{Method} & Modalities & \multicolumn{1}{l}{Gesture Recognition Accurancy} \\ \hline
                     & MAN-DNN                    & sEMG-ACC   & 92.87\%                                           \\
Ninapro DB2          & UAMF-ENAS                  & sEMG-ACC   & 94.10\%                                           \\ 
                     & \textbf{AMF-ENAS}          & sEMG-ACC  & \textbf{ 95.15\% }                                          \\ \hline
                     & MAN-DNN                    & sEMG-ACC   & 90.41\%                                           \\
Ninapro DB3          & UAMF-ENAS                  & sEMG-ACC   & 91.63\%                                           \\ 
                     & \textbf{AMF-ENAS }                  & sEMG-ACC   & \textbf{ 92.50\% }                                          \\ \hline
\multicolumn{1}{c}{} & MAN-DNN                    & sEMG-ACC   & 96.23\%                                           \\
NinaPro DB7          & UAMF-ENAS                  & sEMG-ACC   & 96.71\%                                           \\
                     & \textbf{AMF-ENAS }                  & sEMG-ACC   & \textbf{ 97.19\% }                                          \\ \hline
\end{tabular}%
}
\end{table}

\section{CONCLUSION}
In conclusion, this article introduces ENAS into gesture recognition for the first time and proposes a novel multimodal fusion framework called AMF-ENAS. In gesture recognition, the AMF-ENAS proposes a general multimodal fusion framework that can determine the fusion positions and fusion ratios between modalities. On the NinaPro datasets, the AMF-ENAS successfully fused sEMG and ACC signals, achieving state-of-the-art performance on DB2, DB3, and DB7. In future research, the application of AMF-ENAS could be extended to other modalities, such as images and pressure, broadening its scope beyond the current focus.

\section*{Acknowledgment}
This work was supported in part by the  Natural Science Foundation of China (Grant No. 62276224), in part by the Natural Science Foundation of Hunan Province, China (Grant No. 2022JJ40452), in part by the General Project of Hunan Education Department (Grant No. 21C0077), and in part  by the Scientific Research Foundation of Hunan Education Department (Grant No. 23C0046).


\bibliographystyle{ieeetr}
\bibliography{ref}

\begin{thebibliography}{10}

\bibitem{1}
N.~D. Kahanowich and A.~Sintov, ``Robust classification of grasped objects in intuitive human-robot collaboration using a wearable force-myography device,'' {\em IEEE Robotics and Automation Letters}, vol.~6, no.~2, pp.~1192--1199, 2021.

\bibitem{2}
J.~Wei, Q.~Meng, and A.~Badii, ``Classification of human hand movements using surface emg for myoelectric control,'' in {\em Advances in Computational Intelligence Systems: Contributions Presented at the 16th UK Workshop on Computational Intelligence, September 7--9, 2016, Lancaster, UK}, pp.~331--339, Springer, 2017.

\bibitem{3}
M.~Kim, K.~Kim, and W.~K. Chung, ``Simple and fast compensation of semg interface rotation for robust hand motion recognition,'' {\em IEEE Transactions on Neural Systems and Rehabilitation Engineering}, vol.~26, no.~12, pp.~2397--2406, 2018.

\bibitem{4}
S.~Lee, M.-O. Kim, T.~Kang, J.~Park, and Y.~Choi, ``Knit band sensor for myoelectric control of surface emg-based prosthetic hand,'' {\em IEEE Sensors Journal}, vol.~18, no.~20, pp.~8578--8586, 2018.

\bibitem{guo2021human}
L.~Guo, Z.~Lu, and L.~Yao, ``Human-machine interaction sensing technology based on hand gesture recognition: A review,'' {\em IEEE Transactions on Human-Machine Systems}, vol.~51, no.~4, pp.~300--309, 2021.

\bibitem{5}
M.~Alizadeh-Meghrazi, G.~Sidhu, S.~Jain, M.~Stone, L.~Eskandarian, A.~Toossi, and M.~R. Popovic, ``A mass-producible washable smart garment with embedded textile emg electrodes for control of myoelectric prostheses: A pilot study,'' {\em Sensors}, vol.~22, no.~2, p.~666, 2022.

\bibitem{6}
Y.~Geng, Z.~Yu, Y.~Long, L.~Qin, Z.~Chen, Y.~Li, X.~Guo, and G.~Li, ``A cnn-attention network for continuous estimation of finger kinematics from surface electromyography,'' {\em IEEE Robotics and Automation Letters}, vol.~7, no.~3, pp.~6297--6304, 2022.

\bibitem{11}
L.~Wu, X.~Zhang, K.~Wang, X.~Chen, and X.~Chen, ``Improved high-density myoelectric pattern recognition control against electrode shift using data augmentation and dilated convolutional neural network,'' {\em IEEE Transactions on Neural Systems and Rehabilitation Engineering}, vol.~28, no.~12, pp.~2637--2646, 2020.

\bibitem{12}
E.~Scheme and K.~Englehart, ``Electromyogram pattern recognition for control of powered upper-limb prostheses: state of the art and challenges for clinical use.,'' {\em Journal of Rehabilitation Research \& Development}, vol.~48, no.~6, 2011.

\bibitem{14}
Y.~Xue, Z.~Ju, K.~Xiang, J.~Chen, and H.~Liu, ``Multimodal human hand motion sensing and analysis—a review,'' {\em IEEE Transactions on Cognitive and Developmental Systems}, vol.~11, no.~2, pp.~162--175, 2018.

\bibitem{15}
W.~Wei, Q.~Dai, Y.~Wong, Y.~Hu, M.~Kankanhalli, and W.~Geng, ``Surface-electromyography-based gesture recognition by multi-view deep learning,'' {\em IEEE Transactions on Biomedical Engineering}, vol.~66, no.~10, pp.~2964--2973, 2019.

\bibitem{16}
S.~Duan, L.~Wu, B.~Xue, A.~Liu, R.~Qian, and X.~Chen, ``A hybrid multimodal fusion framework for semg-acc-based hand gesture recognition,'' {\em IEEE Sensors Journal}, vol.~23, no.~3, pp.~2773--2782, 2023.

\bibitem{7}
M.~Ergeneci, K.~Gokcesu, E.~Ertan, and P.~Kosmas, ``An embedded, eight channel, noise canceling, wireless, wearable semg data acquisition system with adaptive muscle contraction detection,'' {\em IEEE Transactions on Biomedical Circuits and Systems}, vol.~12, no.~1, pp.~68--79, 2017.

\bibitem{8}
A.~Phinyomark, P.~Phukpattaranont, and C.~Limsakul, ``Feature reduction and selection for emg signal classification,'' {\em Expert Systems with Applications}, vol.~39, no.~8, pp.~7420--7431, 2012.

\bibitem{lu2014hand}
Z.~Lu, X.~Chen, Q.~Li, X.~Zhang, and P.~Zhou, ``A hand gesture recognition framework and wearable gesture-based interaction prototype for mobile devices,'' {\em IEEE Transactions on Human-Machine Systems}, vol.~44, no.~2, pp.~293--299, 2014.

\bibitem{yang2021dynamic}
Z.~Yang, D.~Jiang, Y.~Sun, B.~Tao, X.~Tong, G.~Jiang, M.~Xu, J.~Yun, Y.~Liu, B.~Chen, {\em et~al.}, ``Dynamic gesture recognition using surface emg signals based on multi-stream residual network,'' {\em Frontiers in Bioengineering and Biotechnology}, vol.~9, p.~779353, 2021.

\bibitem{wei2019surface}
W.~Wei, Q.~Dai, Y.~Wong, Y.~Hu, M.~Kankanhalli, and W.~Geng, ``Surface-electromyography-based gesture recognition by multi-view deep learning,'' {\em IEEE Transactions on Biomedical Engineering}, vol.~66, no.~10, pp.~2964--2973, 2019.

\bibitem{2_0_zheng2022surface}
M.~Zheng, M.~S. Crouch, and M.~S. Eggleston, ``Surface electromyography as a natural human--machine interface: a review,'' {\em IEEE Sensors Journal}, vol.~22, no.~10, pp.~9198--9214, 2022.

\bibitem{2_1_fougner2011resolving}
A.~Fougner, E.~Scheme, A.~D. Chan, K.~Englehart, and {\O}.~Stavdahl, ``Resolving the limb position effect in myoelectric pattern recognition,'' {\em IEEE Transactions on Neural Systems and Rehabilitation Engineering}, vol.~19, no.~6, pp.~644--651, 2011.

\bibitem{2_2_chen2017exploring}
H.~Chen, Y.~Zhang, Z.~Zhang, Y.~Fang, H.~Liu, and C.~Yao, ``Exploring the relation between emg sampling frequency and hand motion recognition accuracy,'' in {\em 2017 IEEE International Conference on Systems, Man, and Cybernetics (SMC)}, pp.~1139--1144, IEEE, 2017.

\bibitem{2_3_englehart1999classification}
K.~Englehart, B.~Hudgins, P.~A. Parker, and M.~Stevenson, ``Classification of the myoelectric signal using time-frequency based representations,'' {\em Medical engineering \& physics}, vol.~21, no.~6-7, pp.~431--438, 1999.

\bibitem{2_4_wu2016wearable}
J.~Wu, L.~Sun, and R.~Jafari, ``A wearable system for recognizing american sign language in real-time using imu and surface emg sensors,'' {\em IEEE journal of biomedical and health informatics}, vol.~20, no.~5, pp.~1281--1290, 2016.

\bibitem{2_5_zhang2011framework}
X.~Zhang, X.~Chen, Y.~Li, V.~Lantz, K.~Wang, and J.~Yang, ``A framework for hand gesture recognition based on accelerometer and emg sensors,'' {\em IEEE Transactions on Systems, Man, and Cybernetics-Part A: Systems and Humans}, vol.~41, no.~6, pp.~1064--1076, 2011.

\bibitem{2_6_kim2008bi}
J.~Kim, J.~Wagner, M.~Rehm, and E.~Andr{\'e}, ``Bi-channel sensor fusion for automatic sign language recognition,'' in {\em 2008 8th IEEE International Conference on Automatic Face \& Gesture Recognition}, pp.~1--6, IEEE, 2008.

\bibitem{2_7_huang2005gaussian}
Y.~Huang, K.~B. Englehart, B.~Hudgins, and A.~D. Chan, ``A gaussian mixture model based classification scheme for myoelectric control of powered upper limb prostheses,'' {\em IEEE Transactions on Biomedical Engineering}, vol.~52, no.~11, pp.~1801--1811, 2005.

\bibitem{2_8_gijsberts2014movement}
A.~Gijsberts, M.~Atzori, C.~Castellini, H.~M{\"u}ller, and B.~Caputo, ``Movement error rate for evaluation of machine learning methods for semg-based hand movement classification,'' {\em IEEE transactions on neural systems and rehabilitation engineering}, vol.~22, no.~4, pp.~735--744, 2014.

\bibitem{2_9_amma2015advancing}
C.~Amma, T.~Krings, J.~B{\"o}er, and T.~Schultz, ``Advancing muscle-computer interfaces with high-density electromyography,'' in {\em Proceedings of the 33rd Annual ACM Conference on Human Factors in Computing Systems}, pp.~929--938, 2015.

\bibitem{2_10_castellini2009surface}
C.~Castellini and P.~Van Der~Smagt, ``Surface emg in advanced hand prosthetics,'' {\em Biological cybernetics}, vol.~100, pp.~35--47, 2009.

\bibitem{2_11_li2021gesture}
W.~Li, P.~Shi, and H.~Yu, ``Gesture recognition using surface electromyography and deep learning for prostheses hand: state-of-the-art, challenges, and future,'' {\em Frontiers in neuroscience}, vol.~15, p.~621885, 2021.

\bibitem{2_12_park2016movement}
K.-H. Park and S.-W. Lee, ``Movement intention decoding based on deep learning for multiuser myoelectric interfaces,'' in {\em 2016 4th international winter conference on brain-computer Interface (BCI)}, pp.~1--2, IEEE, 2016.

\bibitem{2_13_hu2018novel}
Y.~Hu, Y.~Wong, W.~Wei, Y.~Du, M.~Kankanhalli, and W.~Geng, ``A novel attention-based hybrid cnn-rnn architecture for semg-based gesture recognition,'' {\em PloS one}, vol.~13, no.~10, p.~e0206049, 2018.

\bibitem{2_14_he2018surface}
Y.~He, O.~Fukuda, N.~Bu, H.~Okumura, and N.~Yamaguchi, ``Surface emg pattern recognition using long short-term memory combined with multilayer perceptron,'' in {\em 2018 40th annual international conference of the IEEE engineering in medicine and biology society (EMBC)}, pp.~5636--5639, IEEE, 2018.

\bibitem{2_15_betthauser2019stable}
J.~L. Betthauser, J.~T. Krall, R.~R. Kaliki, M.~S. Fifer, and N.~V. Thakor, ``Stable electromyographic sequence prediction during movement transitions using temporal convolutional networks,'' in {\em 2019 9th International IEEE/EMBS Conference on Neural Engineering (NER)}, pp.~1046--1049, IEEE, 2019.

\bibitem{atzori2012building}
M.~Atzori, A.~Gijsberts, S.~Heynen, A.-G.~M. Hager, O.~Deriaz, P.~Van Der~Smagt, C.~Castellini, B.~Caputo, and H.~M{\"u}ller, ``Building the ninapro database: A resource for the biorobotics community,'' in {\em 2012 4th IEEE RAS \& EMBS International Conference on Biomedical Robotics and Biomechatronics (BioRob)}, pp.~1258--1265, IEEE, 2012.

\bibitem{hou2020reformulating}
Z.~Hou, C.~He, and R.~Cheng, ``Reformulating preferences into constraints for evolutionary multi-and many-objective optimization,'' {\em Information Sciences}, vol.~541, pp.~1--15, 2020.

\bibitem{xu2024cluster}
K.~Xu, Y.~Xia, J.~Zou, Z.~Hou, S.~Yang, Y.~Hu, and Y.~Liu, ``A cluster prediction strategy with the induced mutation for dynamic multi-objective optimization,'' {\em Information Sciences}, p.~120193, 2024.

\bibitem{30}
J.~Kennedy and R.~Eberhart, ``Particle swarm optimization,'' in {\em Proceedings of ICNN'95-International Conference on Neural Networks}, vol.~4, pp.~1942--1948, IEEE, 1995.

\bibitem{29}
D.~E. Goldberg, ``Genetic algorithms. pearson education india,'' 2006.

\bibitem{31}
M.~Dorigo, V.~Maniezzo, and A.~Colorni, ``Ant system: optimization by a colony of cooperating agents,'' {\em IEEE Transactions on Systems, Man, and Cybernetics, part b (cybernetics)}, vol.~26, no.~1, pp.~29--41, 1996.

\bibitem{32}
D.~Floreano, P.~D{\"u}rr, and C.~Mattiussi, ``Neuroevolution: from architectures to learning,'' {\em Evolutionary Intelligence}, vol.~1, pp.~47--62, 2008.

\bibitem{33}
E.~Real, S.~Moore, A.~Selle, S.~Saxena, Y.~L. Suematsu, J.~Tan, Q.~V. Le, and A.~Kurakin, ``Large-scale evolution of image classifiers,'' in {\em International Conference on Machine Learning}, pp.~2902--2911, PMLR, 2017.

\bibitem{34}
Y.~Chen, G.~Meng, Q.~Zhang, S.~Xiang, C.~Huang, L.~Mu, and X.~Wang, ``Renas: Reinforced evolutionary neural architecture search,'' in {\em Proceedings of the IEEE/CVF Conference on Computer Vision and Pattern Recognition}, pp.~4787--4796, 2019.

\bibitem{thakkar2018batch}
V.~Thakkar, S.~Tewary, and C.~Chakraborty, ``Batch normalization in convolutional neural networks—a comparative study with cifar-10 data,'' in {\em 2018 Fifth International Conference on Emerging Applications of Information Technology (EAIT)}, pp.~1--5, IEEE, 2018.

\bibitem{kornblith2019better}
S.~Kornblith, J.~Shlens, and Q.~V. Le, ``Do better imagenet models transfer better?,'' in {\em Proceedings of the IEEE/CVF Conference on Computer Vision and Pattern Recognition}, pp.~2661--2671, 2019.

\bibitem{35}
D.~O’Neill, B.~Xue, and M.~Zhang, ``Evolutionary neural architecture search for high-dimensional skip-connection structures on densenet style networks,'' {\em IEEE Transactions on Evolutionary Computation}, vol.~25, no.~6, pp.~1118--1132, 2021.

\bibitem{37}
M.~Shen, K.~Han, C.~Xu, and Y.~Wang, ``Searching for accurate binary neural architectures,'' in {\em Proceedings of the IEEE/CVF International Conference on Computer Vision Workshops}, pp.~0--0, 2019.

\bibitem{38}
M.~G.~B. Calisto and S.~K. Lai-Yuen, ``Self-adaptive 2d-3d ensemble of fully convolutional networks for medical image segmentation,'' in {\em Medical Imaging 2020: Image Processing}, vol.~11313, pp.~459--469, SPIE, 2020.

\bibitem{39}
T.~Cetto, J.~Byrne, X.~Xu, and D.~Moloney, ``Size/accuracy trade-off in convolutional neural networks: An evolutionary approach,'' in {\em Recent Advances in Big Data and Deep Learning: Proceedings of the INNS Big Data and Deep Learning Conference INNSBDDL2019, held at Sestri Levante, Genova, Italy 16-18 April 2019}, pp.~17--26, Springer, 2020.

\bibitem{40}
Z.~Fan, J.~Wei, G.~Zhu, J.~Mo, and W.~Li, ``Evolutionary neural architecture search for retinal vessel segmentation,'' {\em arXiv preprint arXiv:2001.06678}, 2020.

\bibitem{yu2016improved}
F.~Yu, X.~Fu, H.~Li, and G.~Dong, ``Improved roulette wheel selection-based genetic algorithm for tsp,'' in {\em 2016 International Conference on Network and Information Systems for Computers (ICNISC)}, pp.~151--154, IEEE, 2016.

\bibitem{41}
Y.~Du, W.~Jin, W.~Wei, Y.~Hu, and W.~Geng, ``Surface emg-based inter-session gesture recognition enhanced by deep domain adaptation,'' {\em Sensors}, vol.~17, no.~3, p.~458, 2017.

\bibitem{woo2018cbam}
S.~Woo, J.~Park, J.-Y. Lee, and I.~S. Kweon, ``Cbam: Convolutional block attention module,'' in {\em Proceedings of the European Conference on Computer Vision (ECCV)}, pp.~3--19, 2018.

\bibitem{42}
M.~Atzori, A.~Gijsberts, C.~Castellini, B.~Caputo, A.-G.~M. Hager, S.~Elsig, G.~Giatsidis, F.~Bassetto, and H.~M{\"u}ller, ``Electromyography data for non-invasive naturally-controlled robotic hand prostheses,'' {\em Scientific Data}, vol.~1, no.~1, pp.~1--13, 2014.

\bibitem{LDA}
A.~Krasoulis, I.~Kyranou, M.~S. Erden, K.~Nazarpour, and S.~Vijayakumar, ``Improved prosthetic hand control with concurrent use of myoelectric and inertial measurements,'' {\em Journal of Neuroengineering and Rehabilitation}, vol.~14, pp.~1--14, 2017.

\bibitem{gijsberts2014movement}
A.~Gijsberts, M.~Atzori, C.~Castellini, H.~M{\"u}ller, and B.~Caputo, ``Movement error rate for evaluation of machine learning methods for semg-based hand movement classification,'' {\em IEEE Transactions on Neural Systems and Rehabilitation Engineering}, vol.~22, no.~4, pp.~735--744, 2014.

\bibitem{atzori2014classification}
M.~Atzori, A.~Gijsberts, H.~M{\"u}ller, and B.~Caputo, ``Classification of hand movements in amputated subjects by semg and accelerometers,'' in {\em 2014 36th Annual International Conference of the IEEE Engineering in Medicine and Biology Society}, pp.~3545--3549, IEEE, 2014.

\bibitem{liu2016continuous}
J.~Liu, W.~Chen, M.~Li, and X.~Kang, ``Continuous recognition of multifunctional finger and wrist movements in amputee subjects based on semg and accelerometry,'' {\em The Open Biomedical Engineering Journal}, vol.~10, p.~101, 2016.

\bibitem{hudgins1993new}
B.~Hudgins, P.~Parker, and R.~N. Scott, ``A new strategy for multifunction myoelectric control,'' {\em IEEE Transactions on Biomedical Engineering}, vol.~40, no.~1, pp.~82--94, 1993.

\bibitem{krasoulis2019multi}
A.~Krasoulis, S.~Vijayakumar, and K.~Nazarpour, ``Multi-grip classification-based prosthesis control with two emg-imu sensors,'' {\em IEEE Transactions on Neural Systems and Rehabilitation Engineering}, vol.~28, no.~2, pp.~508--518, 2019.

\end{thebibliography}
\end{document}